\RequirePackage{filecontents}

\documentclass{elektr}
\usepackage[all]{xy,xypic}
\usepackage{natbib}
\usepackage[utf8]{inputenc}
\usepackage[english]{babel}
\usepackage[dvipsnames]{xcolor}
\usepackage[pagewise]{lineno}
\usepackage{hyperref}
\usepackage{amsfonts,amssymb,amsgen,amsopn,amsbsy,theorem,graphicx,epsfig}
\usepackage{eufrak,amscd,bezier,latexsym,mathrsfs,eurosym,enumerate}
\usepackage{comment}
\usepackage{subcaption}
\usepackage{adjustbox}
\usepackage[utf8]{inputenc}
\usepackage[english]{babel}
\usepackage{algorithm}
\usepackage{algpseudocode}
\usepackage{courier}
\usepackage[dvipsnames]{xcolor}
\usepackage[pagewise]{lineno}
\usepackage{relsize} 
\usepackage{graphicx} 
\usepackage{placeins}
\usepackage{float}
\usepackage{subcaption}
\usepackage{array,booktabs}
\usepackage{tabularx}
\usepackage[normalem]{ulem}
\usepackage{color}
\usepackage{multirow}
\usepackage{fancyhdr}
\usepackage[symbol]{footmisc} 
\usepackage{amsmath,amssymb,enumitem}
\usepackage{perpage} 
\usepackage{comment}
\usepackage{breqn}
\usepackage{relsize} 
\usepackage{placeins}
\usepackage{breqn}
\usepackage{lineno}
\usepackage{titlesec}
\usepackage{hyperref}
\usepackage{datetime}
\hypersetup{
	colorlinks=true,
	urlcolor=blue,
	citecolor=blue}
\usepackage{cleveref,multirow}

\providecommand{\keywords}[1]
{
	\small	
	\textbf{\textit{Keywords---}} #1
}

\date{\today}
\newcommand*{\bdbackslash}{\textunderscore\kern-.1355ex\textunderscore}
\newcommand*{\cdbackslash}{\_\kern-.1355ex\_}

\title{Semi-supervised Vector-Quantization in Visual SLAM using HGCN\vspace{1cm} \\  \today
}
\author[ Amir Zarringhalam]{
	\textbf{ Amir Zarringhalam$^{1}$\thanks{am.zarringhalam@aut.ac.com}~,Saeed Shiry Ghidary$^{2}$, Ali Mohades Khorasani$^{3}$} \vspace{1cm} \\
	
	$^{1}$Amirkabir University of Technology, Computer Science and  Mathematics, 424 Hafez Ave, Tehran, Iran\\
	$^{2}$Staffordshire University, School of Digital, Technologies and Arts, College Rd, Stoke-on-Trent ST4 2DE, United Kingdom,\\
	Amirkabir University of Technology, Computer Science and  Mathematics, 424  HafezAve, Tehran, Iran\\
	$^{3}$Amirkabir University of Technology, Computer Science and  Mathematics, 424  HafezAve, Tehran, Iran
	\\ [1.8em]
}

\def\E{\ifmmode{\mathbb E}\else{$\mathbb E$}\fi} 
\def\N{\ifmmode{\mathbb N}\else{$\mathbb N$}\fi} 
\def\R{\ifmmode{\mathbb R}\else{$\mathbb R$}\fi} 
\def\Q{\ifmmode{\mathbb Q}\else{$\mathbb Q$}\fi} 
\def\C{\ifmmode{\mathbb C}\else{$\mathbb C$}\fi} 
\def\H{\ifmmode{\mathbb H}\else{$\mathbb H$}\fi} 
\def\Z{\ifmmode{\mathbb Z}\else{$\mathbb Z$}\fi} 
\def\P{\ifmmode{\mathbb P}\else{$\mathbb P$}\fi} 
\def\T{\ifmmode{\mathbb T}\else{$\mathbb T$}\fi} 
\def\SS{\ifmmode{\mathbb S}\else{$\mathbb S$}\fi} 
\def\DD{\ifmmode{\mathbb D}\else{$\mathbb D$}\fi} 

\newcommand{\bse}{\begin{subequations}}
\newcommand{\ese}{\end{subequations}}
\newcommand{\ben}{\begin{enumerate}}
\newcommand{\een}{\end{enumerate}}
\newcommand{\bens}{\begin{enumerate*}}
\newcommand{\eens}{\end{enumerate*}}
\newcommand{\be}{\begin{equation}}
\newcommand{\ee}{\end{equation}}
\newcommand{\bea}{\begin{eqnarray}}
\newcommand{\eea}{\end{eqnarray}}
\newcommand{\baa}{\begin{eqnarray*}}
\newcommand{\eaa}{\end{eqnarray*}}
\newcommand{\bc}{\begin{center}}
\newcommand{\ec}{\end{center}}

\theoremstyle{corollary}

\theoremstyle{lemma}

\theoremstyle{proposition}

\theoremstyle{axiom}

\theoremstyle{conjecture}

\theoremstyle{example}

\theoremstyle{definition}

\theoremstyle{remark}

\begin{document}
	\maketitle



\begin{abstract}
	
	In  this paper, two  semi-supervised appearance based loop closure detection technique, HGCN-FABMAP and  HGCN-BoW are introduced. Furthermore an extension to the current state of the art localization SLAM algorithm, ORB-SLAM, is presented. The proposed HGCN-FABMAP method is implemented in an off-line manner
	incorporating Bayesian probabilistic schema for loop detection decision making. Specifically, we let a Hyperbolic Graph Convolutional Neural Network (HGCN) to operate over the SURF features graph space, and perform vector quantization part of the SLAM procedure.  This part previously  was performed in an unsupervised manner using algorithms like HKmeans, kmeans++,..etc. The main Advantage of using HGCN, is that it scales linearly in number of graph edges. 
	Experimental results shows that HGCN-FABMAP algorithm needs  
	far more cluster centroids than HGCN-ORB, otherwise it fails to detect loop closures. Therefore we consider HGCN-ORB to be more efficient in terms of memory consumption, also we conclude the superiority of HGCN-BoW and HGCN-FABMAP with respect to other algorithms. \\ \\ 
	\keywords{	HGCN(Hyperbolic Graph Convolutional Neural Network), FABMAP2, ORB-SLAM, Vector Quantization
	}
\end{abstract}

\section{Introduction}

Mobile autonomous robots that  can navigate in long period of time in troublesome circumstances can assist humans.
In many situations robots doesn't have any previous knowledge from their surroundings, and there is no Global Positioning System (GPS) available for the sake of assistance. Therefore, robots might not have localized themselves with sufficient accuracy. The navigation problem in unknown environment without GPS assistance is known as 
simultaneous localization and mapping (SLAM).\\
In visual SLAM, place recognition is  known as loop closure detection problem.
Although  GPS allows devices to navigate in outdoor environments with sufficient localization accuracy, there are still many situations like indoor places, tunnels and even urban places where skyscrapers mask the GPS signals, so GPS might not  work reliably.\\
As the main parts of the SLAM process, loop closure detection and localization have preoccupied many researchers, and they are still  open problems. Approaches like Fast Appearance-Based Mapping (FAB-MAP) had a magnificent success in  several long range experiments like 1000 km trajectory planing in England \cite{5509547}, Still, this method suffers from low recall.\\
Our task here is to increase accuracy and  recall of FABMAP2 trough an alternative vector quantization (VQ) process. 
 We conjecture that for large scale SLAM, accuracy degradation might be due to applying  improper clustering on the SURF feature space. As the number of features in the sequence of images increase, they form a tree like structure with too many nodes overlapping each other (at the bottom of the tree). If we map these features to a Poincare ball, features will  become well  separated, and a more clusterable  data point we get as the result of mapping.\\ 
It is a common practice for long range SLAM to use BoW schema. BoW technique has originally been  used  in text retrieval context. However, currently this method is used  extensively in computer vision communities.
Computational complexity of BoW based SLAM will not increase significantly with respect to other methods, and this method has been used  as a standard method in loop closure detection module of the SLAM algorithm.  One of the main problem of  BoW based SLAM in large scale is that the vocabulary size will get large since map is constructed from numerous images, therefore  
VQ process in which  thousands of features should be matched against each other will be a lot more expensive with respect to small scale SLAM. We are also confronting  a low Recall which corresponds to handling perceptual aliasing.  \\

In this paper, our primary contribution is that we rebuild the vocabulary and replace the BoW construction block, which is usually obtained from unsupervised algorithms like Kmeans++, with a semi-supervised method HGCN. Next, we fuse this model into two SLAM alorithm ORB-SLAM and FABMAP2. Results show the superiority of the new FABMAP algorithm  with respect to original FABAMAP2.
This paper is structured  as follow, in section 2 a review of the literature is presented,  in section 3 a schematic overview of our main contribution is presented, section 4 is dedicated to Experiments and results and in section 5 the future direction of this research is given.\\

\section{Related works}

Most of the works in BoW SLAM use unsupervised approaches for loop closure detection, we are not aware of any semi-supervised or supervised approaches, tackling this problem.
In \cite{Hajebi2011FastAN,DBLP:journals/corr/HajebiZ13} and \cite {Khan2015IBuILDIB}, K.Hajebi introduced a graph based nearest neighbor search method, stochastic graph nearest neighbor (SGNN), to speed up BoW image retrieval. SGNN starts from a random node in the graph, and it performs a greedy
search on the graph structure. In SGNN algorithm, the computational cost of finding  an approximate nearest neighbor for a random query node Q is  bounded by $min\{n^d, 2n^{\frac{1}{d}}d^2\}$ where $n $ is the number of points and  $d$ is the dimension of  hypercube of volume 1 in which the points are randomly drawn from. Still, in this algorithm there is a distance constraint between each node and its nearest neighbor.\\

In \cite{KEJRIWAL201655}, Nishant Kejriwal proposed a bag of word pairs (BoWP) approach which uses spatial co-occurrence of words to overcome usual bag of words techniques limitations such as perceptual aliasing. This algorithm is implemented in an on-line fashion. \\

In \cite{Khan2015IBuILDIB}, an online incremental method based on binary visual vocabulary is employed for loop closure detection problem. The binary vocabulary generation process is based on tracking features across consecutive images, which results in being invariant to the robot pose and make it suitable for loop closure detection. Also, it represents 
a schema for updating and generating binary visual words, coupled with a simplified likelihood function for LCD problem.\\



In PTAM, an alternative approach for current SLAM solutions is employed for tracking and mapping unknown environment.
That is, instead of restricting  ourselves to frame by frame scalability of incremental mapping, which results a sparse  map with high quality features, this paper proposes another solution having far more dense features but with lower quality features. Briefly, this method presents a  camera pose detection in an unknown environment.  Although,  this approach has been previously addressed several SLAM solutions, PTAM specifically  focuses on handling  involuntary 
camera movements, in small augmented reality environments. \\
 
In \cite{DBLP:journals/corr/abs-1802-05909}, a novel appearance based loop closure detection method,   iBoW-LCD, is presented.  This method make use of incremental BoW schema on descriptors, therefore it doesn't need any training vocabulary. Furthermore, iBoW-LCD is built on a effective mechanism to group similar images close in time complexity associated with binary feature words. For a detailed explanation about this algorithm see \cite{DBLP:journals/corr/abs-1802-05909}.\\

In \cite{fer2021ov2slam}, $OV^{2}$ is  presented as an online visual SLAM algorithm for both monocular as well as stereo cameras. $OV^{2}$ is accurately designed to perform 4 tasks, tracking, mapping, Bundle adjustment and loop closing in an online fashion using a multi-threaded  system. Tracking can indirectly be obtained using Lucas-Kanade optical-flow formulation. This module provide us camera pose estimation with respect to camera's frame rate. In SLAM process the mapping module works according to key frame rate. This module guarantees that continuous localization can be  
obtained through populating the 3D map and drift minimization in a single local map tracking step.
Loop closing is performed trough online BoW method, iBoW-LCD. This method totally preserve the model performance without  neglecting the accuracy degeneration. \\

In  \cite{5509547}, FABMAP2,  a probabilistic approach in which every environment is made up of several discrete locations is presented. Each of these locations is associated with an appearance model, and it specifically
addresses the following problem: providing  two image sequence with similar context (Train and Test) datasets, it determines whether the the test sequence contains any loop. Though, this is a difficult problem due to the different lightning condition that each image sequence is having and different view point and  dynamicity of the  world,..,etc. Therefore a system beyond simple image similarity computation is needed. FABMAP2 addresses this problem through learning probability distribution over the space of images.\\

For sequence of observations extracted from images taken by  the robot, we either update our belief about the locations present in the map or a new location is created in the map. If the current scene element matches the location with probability higher than  
$0.999$ this element is associated with the with the location otherwise it is associated with a completely new location.
The reason behind considering a reasonably high threshold  can be justified as follow:
Incorrectly assigning two different observations to the same place has far more negative effects than incorrectly assigning the observation to a new map point, because if we assume that two completely different observation are  originated from same place, the appearance model derived from this assumption, tends toward average appearance. Furthermore, as the new observations are similar to the average location with a high probability, and as the robot collects new images from the path  belonging to the average location, the probability of this place being a loop closure, gets higher. As this process repeats, the filter breaks immediately. In contrast, incorrectly considering a new observation as a new map point, results in creating two different location in the map. When an observation belonging to this location is received  by the robot the probability of being a loop closure for this location will be divided by two. Therefore the loop closure detection error will not increase too much.

One of the main advantages of FABMAP2  is that it incorporates the correlation between visual features which improves the system performance. Two approximation inference procedures for this model are defined using Bennet's inequality and the inverted index technique. Using these inferences FABMAP2 can be applied to large scale navigation problems. For a detailed mathematical explanation about this method refer to [\cite{5613942} - \cite{williams08iros}].\\

As another example, ORB-SLAM which is based on the visual SLAM, was firstly  introduced in 2015 by Raul Mur{-}Artal. In 2017 some improvement was made to the algorithm in a way that it was capable of handling stereo cameras as well as cameras equipped with depth detection. ORB-SLAM has been built on main ideas of PTAM \cite{4538852}, parallel tracking and mapping, a work by Klein and Murry. However there are several drawbacks associated with PTAM method, first,  this algorithm is only applicable in small scale environments, although it uses a very efficient key frame selection method. Improper loop closing method and being in need of human interaction for map bootstrap are among other limitations of this algorithm. Meanwhile ORB-SLAM as a monocular SLAM method has overcome the associated problem with PTAM's algorithm and has the following advantages:
\begin{enumerate}
\item  For all tasks, tracking, mapping, re-localization and loop closing, it uses an identical set of  features. Through using ORB features a real time performance even without using GPU's is achievable, \cite{DBLP:journals/corr/Mur-ArtalMT15}.
\item Using the co-visibility graph, tracking, which is responsible for new key frame insertion, localization of camera associated with every frame and mapping are performed in a local co-visible area. Therefore  large scale navigation is possible,  \cite{DBLP:journals/corr/Mur-ArtalMT15}.

\item Camera re-localization, robust to view point and illumination, can be achieved in real time.
\item Using the survival of the fittest approach for the key frame insertion, which is very cautions at discarding futile key frames and very gracious at key frame insertion, robot is able operate long-term.
\end{enumerate} 

\section{Methods}
In this paper, we assume that SURF features extracted from sequence of images form a graph or a tree with too many leafs. It's a common practice to embed hierarchical trees, like graphs in a hyperbolic space, which is a Reimanian manifold with constant negative curvature $\frac{-1}{k}$, where $\sqrt{k}$ is the radius of the Poincaré ball. Hyperbolic geometry differs from euclidean geometry by the parallel postulate, i.e in hyperbolic spaces there are infinite number of lines parallel to the given line through the given point.

\subsection{HGCN}
GCNs (Graph Convolutional Neural Networks) embed graph of instances into the points in Euclidean space. This solves many problems like large distortion when embedding a real world graphs. In addition, hyperbolic geometry enables graph embedding with small distortion, however there are several drawback associated when using hyperbolic geometry. Issues like lack of convolutional and aggregation operators in hyperbolic spaces. In  \cite{DBLP:journals/corr/abs-1910-12933}, these operators are defined for hyperbolic spaces suitably.
The message passing mechanism can be done in three steps in HGCNs. At each layer, first a  hyperbolic feature transformation is done, next neighbour’s embedding aggregation is performed in the tangent space of the center node and  finally HGCN  projects the result back to a hyperbolic space with different curvature. These three steps are formulated as follows:

\begin{equation}
h_{i}^{l,H} = (W^{l}\otimes^{K_{l-1}} x_{i}^{l-1,H}) \oplus ^{K_{l-1}}b_{l}  \\
\end{equation}
\begin{equation}
y_{i}^{l,H} = AGG^{k_{l-1}}h^{l,H}_{i}  \\
\end{equation}

\begin{equation}
x_{i}^{l,H} = \sigma^{\otimes^{k_{l-1},k_{l}}} y_i^{(l,H)} 
\end{equation}
\subsection{HGCN-Vector Quanization}
In the context of computer vision,  vector quantization (VQ) is a procedure trough which features of an images 
are assigned to their nearest corresponding cluster centroids. That is, we first cluster the the extracted SURF feature space using a hyperbolic graph convolutional neural network. Next, using the obtained clustering centroids, we calculate the quantized version of the vectors as shown below: \\
\begin{equation}
Q_{k}^{(i)} = \{ \underline{x} : || \underline{x}-\underline{\hat{x}}_{k}^{i} ||  \leqslant  ||\underline{x}-\underline{\hat{x}}_{k}^{j}|| ; \forall j \in \{1,...,n\} \} 
\end{equation}

\subsection{HGCN-FABMAP}
As part of every BoW based image retrieval task, we need to construct BoW representation of the images.
This can be achieved through constructing clustering centroids from extracted SURF/ORB features of images.
In the HGCN-FABMAP method a hyperbolic graph convolutional neural network is used to  construct the clustering  centroids, so the vector quantization part is done semi-supervisedly. One of the main advantages of using HGCN over  classical unsupervised methods is that it scales linearly in number of extracted features.

HGCN-FABMAP is build upon FABMAP2. In the initial version of FABMAP (FABMAP1) which is described in the appendix section, locations are not constrained, and the log-likelihood of each location can take any value. However, in FABMAP2, various restrictions are put on the likelihood values of places.  That is the likelihoods of the locations where a visual word q has never been seen before,  are constrained to have a fixed value. Furthermore,   
the likelihoods can be incremented, using one of the following values for a given word and a particular location:

\begin{equation}
	\label{noisy}
	\left\{ \begin{array}{*{35}{l}}
		Cas{{e}_{1}}:({{S}_{q}}=1,{{S}_{{{p}_{q}}}}=1)  \\
		Cas{{e}_{2}}:({{S}_{q}}=1,{{S}_{{{p}_{q}}}}=0)  \\
		Cas{{e}_{3}}:({{S}_{q}}=0,{{S}_{{{p}_{q}}}}=1)  \\
		Cas{{e}_{4}}:({{S}_{q}}=0,{{S}_{{{p}_{q}}}}=0)  \\
	\end{array} \right.
\end{equation}

As discussed in the appendix section, $Z_{k} = \{z_{1},...,z_{|\nu|}\}$ signifies a local scene observation at time k, with $z_i$ being a binary variable denoting the presence or absence of the $i^{th}$ word in the vocabulary, also each $z_i$ is a  noisy measurements of every visual word $q$.  The states in formula \ref{noisy} depend on the presence and absence of visual word $q$ and it's parent $p_q$. Since the  observations are typically 
sparse and as we are using the Inverted-Index technique, Case4, in which  both   $ z_q $ and $ z_{p_q} $ are zero, is the most probable scenario for our observations, therefore the default likelihood  for every location $L_{i}$ is computed using Case4 as shown in algorithm \ref{loop}.

\begin{algorithm}[!htp] 
\caption{Default likelihood}
\label{loop}
\begin{algorithmic}[1]
	\For{q $ \leftarrow $ 1 to $\#clusters$}
	\State Locations $\leftarrow$ Inverted-Index[q]
	\For{$L_i$ in Locations}
	\State Log-Likelihood[$L_i$] += $d_1$
	\EndFor
	\EndFor
\end{algorithmic}
\end{algorithm}	

The HGCN-FABMAP method, like the FABMAP2 algorithm, returns a confusion matrix that can be used to identify loop closures.
 
As part of constructing the confusion matrix we need to initialize the default log-likelihood of locations. Algorithm \ref{loop} is used for the sake of this purpose. The default likelihoods are calculated once, when the location is added to the map.
In the  formula \ref{equ1} for every image, $q$ is the index of the word in vocabulary, and it ranges from $0$ to the $|\mathcal{C}|$.

\begin{flalign}
	\label{equ1}
	d_1^* = \frac{(1-CLtree(1,q)) \times 0.61 \times (1-CLtree(3,q))}{CLtree(1,q) \times 0.39 \times (1-CLtree(3,q)) + (1-CLtree(1,q)) \times 0.61 \times (1-CLtree(3,q))} &&
\end{flalign}
\begin{flalign}
	d_1 = \log(d_1^*) &&
\end{flalign}
While processing new observations, we only need to adjust the default likelihood of locations. This adjustment is represented in algorithm  \ref{loop1}. In the first part of the algorithm we only consider the adjustments in which $ z_q = 1 $, and weights will be updated according to the content of current observation using the terms $d_4$ and $d_3$. The last portion of the algorithm  uses $d_2$ to adjust the likelihood, where word $q$ is not observed but it's parent in Chow-Liu tree is present. \\

\begin{algorithm}[!hp] 
\caption{Filling Log-likelihood}
\label{loop1}
\begin{algorithmic}[1]
	\For{$ z_q \in Z$ where $z_q=1$}
	\State Locations $\leftarrow$ Inverted-Index[q]
	\For{$L_i$ in Locations}
	\If {CLTree(0,q) $ \> 0 $}
	\State Log-Likelihood[$L_i$] += $d_4 - q_1$
	\Else
	\State Log-Likelihood[$L_i$] += $d_3 - q_1$
	\EndIf
	\EndFor
	\EndFor
	\For{$ z_q \in Z$ where $z_q=0$ and $z_{p_{q}}=1$}
	\State Locations $\leftarrow$ Inverted-Index[q] 
	\For{$L_i$ in Locations}
	\State Log-Likelihood[$L_i$] += $d_2 - q_1$
	\EndFor
	\EndFor
\end{algorithmic}
\end{algorithm}

\begin{flalign}	
\label{equ7}
\mathlarger{d_{2_{num}}} = \frac{CLtree(1,q) \times 0.61 \times (1-CLtree(2,q))}{(1-CLtree(1,q)) \times 0.39 \times CLtree(2,q)} &&
\end{flalign}
\begin{flalign}
\mathlarger{d_2} =\mathlarger{ \log(\frac{d_{2_{num}}}{1- d_{2_{den}}})} - \mathlarger{d_q} &&
\end{flalign}
\begin{flalign}
\mathlarger{d_{2_{den}}} =\frac{CLtree(1,q) \times 0.61 \times CLtree(2,q)^{2} \times 0.39}{[(1-CLtree(1,q)) \times 0.61 \times (1-CLtree(2,q)) + CLtree(1,q) \times 0.39 \times CLtree(2,q)] \times (1- 0.39\times CLtree(1,q)) }  &&
\end{flalign}
\begin{flalign}
\mathlarger{d_{3_{den}}}= \frac{0.61 \times CLtree(1,q) \times (1-CLtree(1,q)) \times 0.39 \times CLtree(3,q) }{(1-CLtree(1,q)) \times CLtree(1,q) \times 0.61 \times (1-CLtree(3,q))} &&
\end{flalign}

\begin{flalign}
\mathlarger{d_{3_{num}}} = \frac{(1-CLtree(1,q)) \times 0.39 \times CLtree(3,q)}{(1-CLtree(1,q)) \times 0.39 \times CLtree(3,q) + CLtree(1,q) \times 0.61 \times (1-CLtree(3,q))} &&
\end{flalign}

\begin{flalign}
\mathlarger{d_3} = \mathlarger{\log(\frac{d_{3_{num}}}{d_{3_{den}}})} - \mathlarger{d_q} &&
\end{flalign}

\begin{flalign}
\mathlarger{d_4^* }= \frac{CLtree(1,q) \times 0.61}{1- 0.39 \times CLtree(1,q)} &&
\end{flalign}

\begin{flalign}
\label{equ16}
\mathlarger{d_4} = \mathlarger{\log(d_4^*)} - \mathlarger{d_q} &&
\end{flalign}
  In equations \ref{equ7} - \ref{equ16} CLtree refers to Chow-Liu tree obtained from the training vocabulary using the BoW representation which is the result of applying HGCN on the images features space. For more information about the derived formulation, refer to the source code in \citep{5509547}.
\begin{figure*}[!htp]
\centering
\includegraphics[scale = 0.62]{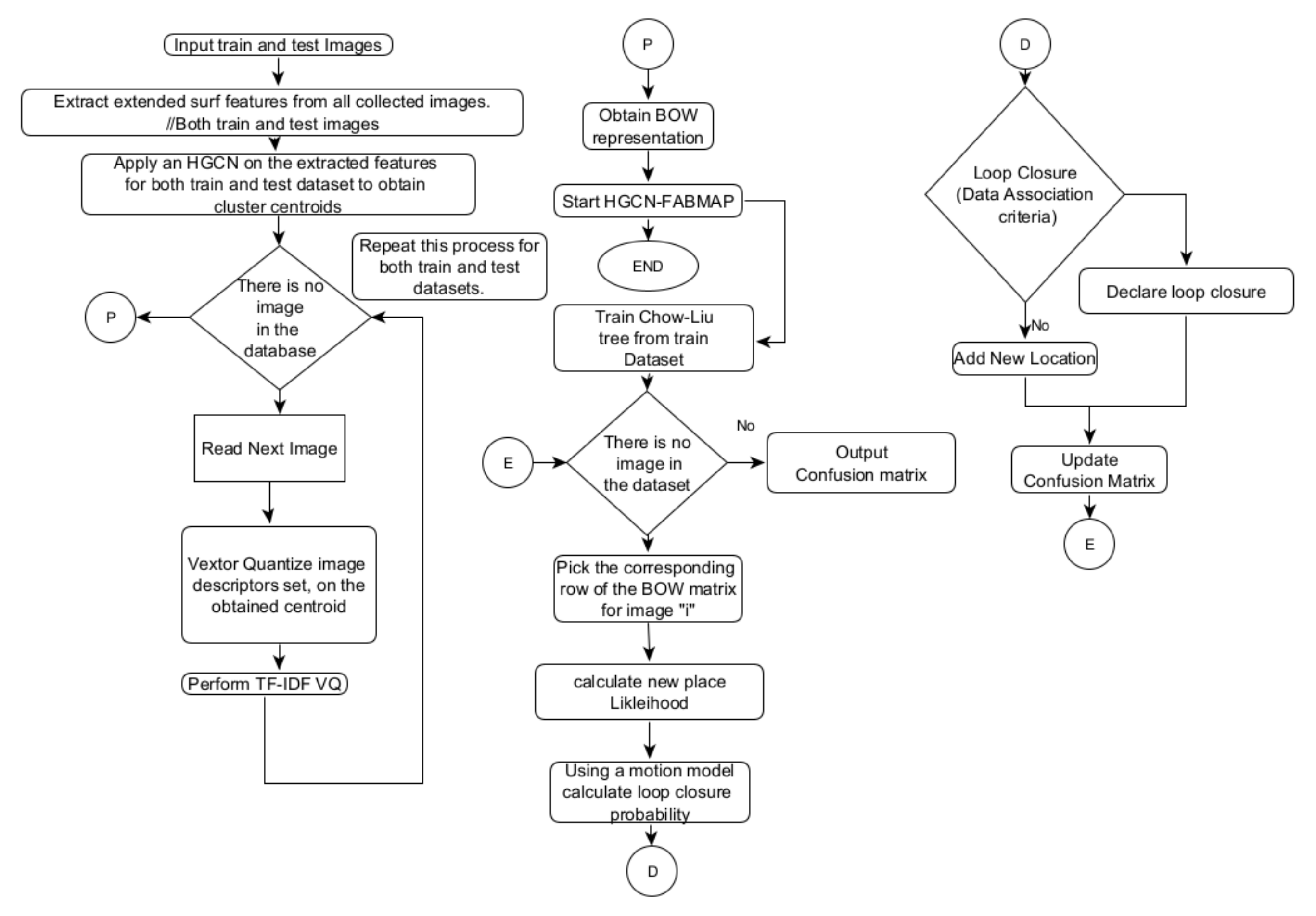}
\caption{ HGCN-FABMAP}\label{b}
\end{figure*}

For the introduced methods in this paper, either loop closure detection or mapping the environment, 
first we extract set of features (ORB/extended 128-D SURF) from all  images in the database. Due to high memory consumption of applying HGCN,  the dimension of the feature matrix is reduced to 20 using PCA so that the features matrix remains meaningful \cite{6782160}. As part of applying HGCN we need to calculate $N$ connected nodes to every feature in the space, considering features form a graph.
To achieve this we have used the parallel Faiss \cite{JDH17} algorithm, otherwise this task would be intractable on personal computers. For the modified FABMAP2 method, after extracting features for both training and testing data, BoW representation of images is calculated. As it is shown in  figure \ref{b}, first for every image the set of all features is obtained and vector quantized on the cluster centroids of the features extracted from all train/test images. Next TF-IDF weighting is applied on the BoW representation of the image. Repeating this procedure for every image results in BoW representation of all images. Next, only for training data a Chow-Liu tree is trained using the cluster centroids and BoW representation. Using the obtained CLtree and BoW representation of image $i$, we determine whether a new place is detected or we have encountered a loop closure. Through applying a motion model and considering the threshold criteria discussed in 2.1. At the end of the algorithm confusion matrix is outputted.

\subsection{HGCN-ORB SLAM} 

In this method we assume that ORB features extracted from sequence of images form a graph or a tree with too many leafs. In off-line phase of ORB-HGCN SLAM, for the sake of place recognition and loop closure detection a new BoW representation of all images is firstly created through applying a HGCN over the extracted features. This BoW representation is then used in the on-line phase. The main step of this algorithm is as follow: \\
Through receiving the sequence of images, ORB features are extracted from each image, and by comparing features of the current frame with several previous frames the current camera position and robot pose is detected.
The features are extracted in a way that they are robust to changes in view's angle and camera orientation. \\

\section{Experimental results}

The experiments are performed on both, indoor and outdoor datasets. Four  methods BoW-SLAM, HGCN-BoW, FABMAP2 and HGCN-FABMAP are applied on Lip6 indoor dataset, TUM sequence 11, Stlucia (train/test), Newer college(short experiment )and New College dataset. For ORB-SLAM  the experiment is performed on TUM  Freiburg3  long office household. In the following section we briefly review the datasets used in this article. To implement HGCN-FABMAP we have modified the C++ package provided by \cite{Glover}.
\newline

\subsection{Datasets}

\subsubsection{Datasets Used in HGCN-FABMAP, HGCN-BoW }

\hspace{0.9cm} \textbf{Lip6indoor dataset} :  Contains 350 images from a lab environment with a narrow corridor\footnote[2]{https//animatlab.lip6.fr/AngeliVideosEn}.\\

\textbf{TUM sequence 11}  : Contains 1500 images from a lab environment with different lighting condition \footnote[3]{https://vision.in.tum.de/data/datasets/mono-dataset}.\\

\textbf{New College [Cummins and Newman 2008]}:This dataset  has been taken from the Oxford university campus which includes complex repetitive structures. This is an stereo dataset with left and right sequence. We have used the left sequence containing 1073  with resolution of 640×480 as our training dataset for newer college dataset.\\

\textbf{Newer College}: This dataset is a two minute movie and it contains 3 loops. We have converted it to 
200 images (Although this is not recommended for FAMMAP2 method). In these dataset camera is experiencing cluttered movements.\\

\textbf{Stlucia suburbs}: This dataset two training and testing movie. Training data converted it to 
540 images, and test dataset is converted to approximately 1000 images.\\
\subsubsection{Datasets Used HGCN-ORB}	
\textbf{Freiburg$3$}\footnote[4]{https://vision.in.tum.de/rgbd/dataset/freiburg3/}
: This indoor dataset contains 2500 images taken from surrounding area of a table 
and it contains one loop.\\ 
\begin{figure*}[!htp]
\centering
\includegraphics[scale = 0.85]{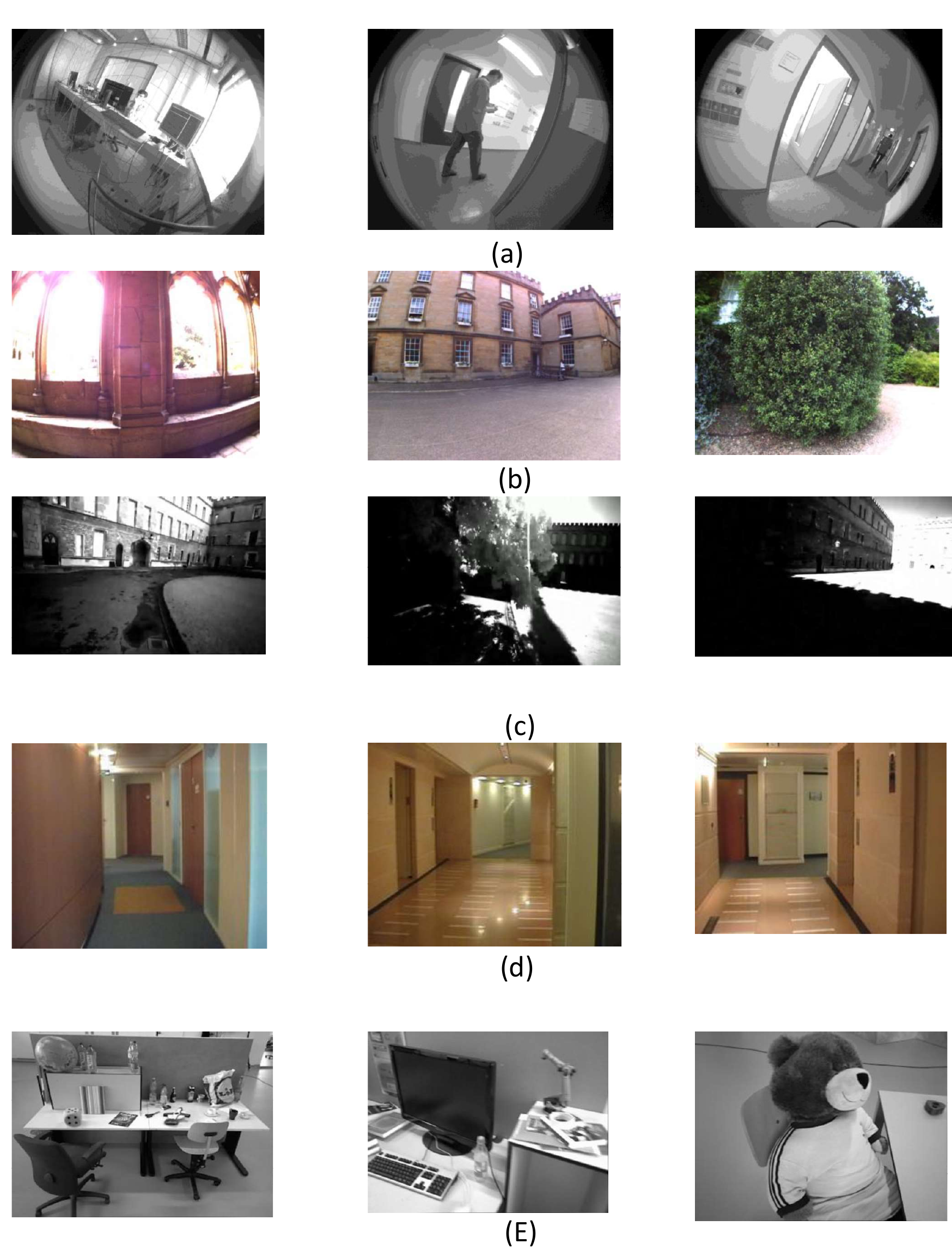} \\
\caption{ Example images of the used datasets. (a) TUM sequence $11^{nt}$ (b) New College (c) Newer College (d) Lip6Indoor (e) Frieburg3 datasets}\label{c}
\end{figure*}

\subsection{Loop closure detection Performance Metrics}

In this article we have borrowed the metrics used in \cite{zhong}. That is recall and accuracy of loop closure detection are defined as follow:

\begin{equation}
\label{equation9}
recall  =\mathlarger{\frac{\sum\limits_{i} \sum\limits_{j}{((Confusion \hspace{0.1cm} Matrix[i][j] \hspace{0.1cm} > threshold)\hspace{0.1cm} \wedge ground \hspace{0.1cm} truth[i][j]==1) }}{\sum\limits_{i} \sum\limits_{j}{ground \hspace{0.1cm} truth[i][j]==1}}}
\end{equation}

Each entry $(i^{th},j^{th})$ of the ground truth matrix  indicates whether images $i$ and $j$ were taken from the same location, if so, the entry is 1, otherwise it is zero.
True positives, the numerator of formula ~\ref{equation9} and ~\ref{equation10}, is the number of elements in the confusion matrix greater than a given threshold, if their corresponding  elements in the ground truth matrix are also ones.
\\

Accuracy is defined as below:\\

\begin{equation}
\label{equation10}
accuracy  =\mathlarger{\frac{\sum\limits_{i} \sum\limits_{j}{((Confusion \hspace{0.1cm} Matrix[i][j] \hspace{0.1cm} > threshold)\hspace{0.1cm} \wedge ground \hspace{0.1cm} truth[i][j]==1) }}{\sum\limits_{i} \sum\limits_{j}{(Confusion \hspace{0.1cm} Matrix[i][j]==1  \hspace{0.1cm} > threshold)}}}
\end{equation}

\subsection{Disscusion}
Method FABMAP2 and it's HGCN extended version require training and testing  datasets not to have any   repetitive path. It is also necessary that the two datasets  share  similar context, otherwise the algorithm does not perform well. For  all the algorithms tested in this article we have extracted the standard 128-d surf features from the dataset. Next PCA (with number of principal component 20) is 
applied on feature matrix, and we feed the matrix to the Faiss algorithm \cite{JDH17} to construct the adjacency graph. Number of closest node to every other node is set to 9 and experiments shows it suffices. Next we feed the obtained data to a 2 layer hyperbolic graph convolutional neural network, which is implemented in pythorch \cite{DBLP:journals/corr/abs-1910-12933} to obtain the labels, BoW representation is obtained afterward.\\

\begin{figure*}[!htp]
\centering
\includegraphics[scale=0.4]{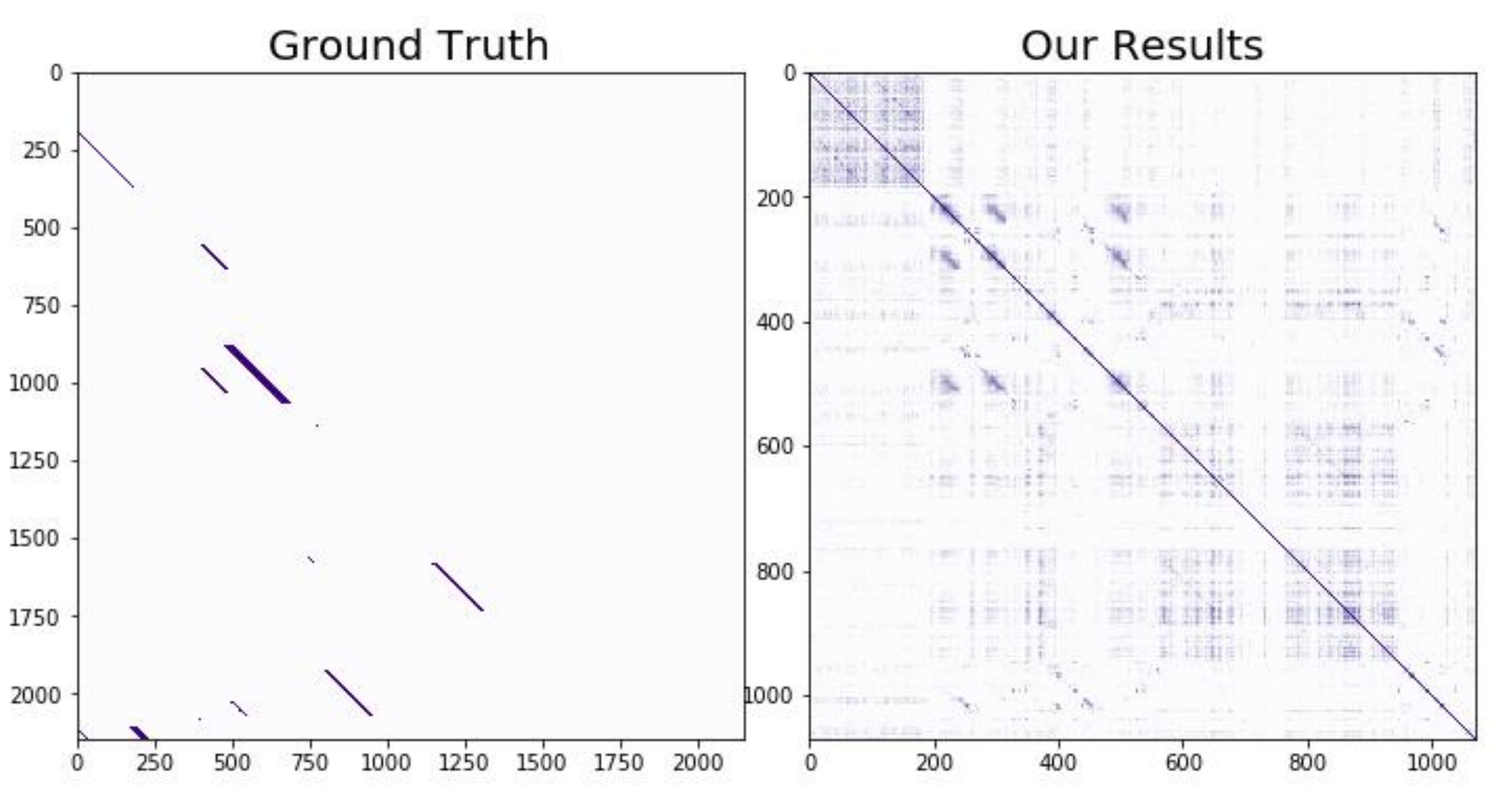}
\caption{ Ground truth (left) and HGCN-BoW (right) confusion matrices for New College dataset}\label{d}
\end{figure*}

First row of the table \ref{tab 1} shows that HGCN-FABMAP and HGCN-BoW have outperformed the regular BoW and FABMAP2 algorithms, for the New College  dataset considering Stlucia(train) dataset to be our training data. Confusion matrix is also plotted for FABMAP2 and HGCN-BOW at Fig \ref{d} for  New College dataset.
The reason behind selecting HGCN-BOW is that, we believe the image comparison have a fuzzy nature instead of being binary.\\ \\

Each element of the confusion matrix plotted in Figure \ref{d}, is the dot product between two rows of HGCN-BoW representation of New College dataset.

\begin{table}
\centering
\refstepcounter{table}
\caption*{ \textbf{Table 1: } Comparing HGCN-FABMAP and FABMAP2 methods for LCD.}
\label{tab 1}
\begin{tabular}{|l|l|l|l|l|l|} 
	\hline
	Train Dataset                                                 & Test Dataset                                                                              & \multicolumn{2}{l|}{~HGCN-FABMAP}                                                                                                                                                      & \multicolumn{2}{l|}{~ ~ FABMAP2}                                                                                                                    \\ 
	\hline
	&                                                                                           & ~ \%Acc                                                                                 & \%Rec                                                                                        & ~Acc~ ~ ~                                                                 & ReC                                                                     \\ 
	\hline
	\begin{tabular}[c]{@{}l@{}}~Stlucia~ ~~\\~311417\end{tabular} & \begin{tabular}[c]{@{}l@{}}New College\textcolor[rgb]{0.502,0,0}{~}~\\160500\end{tabular} & \begin{tabular}[c]{@{}l@{}}\textbf{\%50.07} \\ \%56 \\ \textbf{ \%77.3}\end{tabular}    & \begin{tabular}[c]{@{}l@{}}\textbf{\%100}\\\%86\\\textbf{ \%68.62}\end{tabular}              & \begin{tabular}[c]{@{}l@{}}\%50.02\\\%56.8\\\%77.6\end{tabular}           & \begin{tabular}[c]{@{}l@{}}\%100\\\%82\\\%63\end{tabular}               \\ 
	\hline
	\begin{tabular}[c]{@{}l@{}}TUM seq11~\\3924499\\\end{tabular} & \begin{tabular}[c]{@{}l@{}}Lip6Indoor~ ~ ~\\27189\end{tabular}                            & \begin{tabular}[c]{@{}l@{}}\%50.3\\\textbf{\%54.9}\\\%59.55\end{tabular}                & \begin{tabular}[c]{@{}l@{}}\%100\\\textbf{\%90.5}\\\%81.5\end{tabular}                       & \begin{tabular}[c]{@{}l@{}}\textbf{\%50.34}\\\%54.9\\\%52.59\end{tabular} & \begin{tabular}[c]{@{}l@{}}\textbf{\%100}\\\%90.5\\\%81.2\end{tabular}  \\ 
	\hline
	\begin{tabular}[c]{@{}l@{}}Stlucia\\311417\end{tabular}       & \begin{tabular}[c]{@{}l@{}}~ Stlucia\\~ 235510\end{tabular}                               & \begin{tabular}[c]{@{}l@{}}\textbf{\%75}\\\textbf{\%54.05}\end{tabular}                 & \begin{tabular}[c]{@{}l@{}}\\\textbf{\%72}\\\textbf{\%95.85}\\\textbf{}\end{tabular}         & \begin{tabular}[c]{@{}l@{}}\%50.8\\\%50.4\\\%59.55\end{tabular}           & \begin{tabular}[c]{@{}l@{}}\%100\\\%90\\\%81.5\end{tabular}             \\ 
	\hline
	\begin{tabular}[c]{@{}l@{}}New College\\160500\end{tabular}   & \begin{tabular}[c]{@{}l@{}}Newer College\\194000\end{tabular}                             & \begin{tabular}[c]{@{}l@{}}\textbf{\%90.2}\\\textbf{\%97.4}\\\textbf{\%99}\end{tabular} & \begin{tabular}[c]{@{}l@{}}\textbf{\%80.85}\\\textbf{\%72.34}\\\textbf{\%67.75}\end{tabular} & \begin{tabular}[c]{@{}l@{}}\%71\\\%79\\\%80.02\end{tabular}               & \begin{tabular}[c]{@{}l@{}}\%80.3\\\%72.34\\\%67.55\end{tabular}        \\
	\hline
\end{tabular}
\end{table}

\begin{table}
\centering
\refstepcounter{table}
\caption*{\textbf{Table 2: }Comparing HGCN-BOW and BoW methods for LCD.}
\label{tab 1}
\begin{tabular}{|l|l|l|l|l|l|} 
	\hline
	Train Dataset~                                                & Test Dataset                                                                               & \multicolumn{2}{l|}{~ ~HGCN-BoW}                                                                                                                                                  & \multicolumn{2}{l|}{~ ~ ~BoW~ ~}  \\ 
	\hline
	&                                                                                            & ~Acc                                                                                       & Rec                                                                                  & Acc   & Rec                       \\ 
	\hline
	\begin{tabular}[c]{@{}l@{}}~Stlucia~ ~~\\~311417\end{tabular} & \begin{tabular}[c]{@{}l@{}}New College~\textcolor[rgb]{0.502,0,0}{~}~\\160500\end{tabular} & \begin{tabular}[c]{@{}l@{}}\textbf{\%50.07}\\\textbf{\%91}\end{tabular}                    & \begin{tabular}[c]{@{}l@{}}\textbf{\%100}\\\textbf{\%70.5}\end{tabular}              & \%50  & \%85                      \\ 
	\hline
	\begin{tabular}[c]{@{}l@{}}TUM seq11~~\\3924499\end{tabular}  & \begin{tabular}[c]{@{}l@{}}~ Lip6Indoor~ ~ ~~\\~ 27189\end{tabular}                        & \begin{tabular}[c]{@{}l@{}}\textbf{\%50.037}\\\textbf{\%54.48}\\\textbf{\%60}\end{tabular} & \begin{tabular}[c]{@{}l@{}}\textbf{\%100}\\\textbf{\%90}\\\textbf{\%81}\end{tabular} & \%34  & ~\%100                    \\ 
	\hline
	\begin{tabular}[c]{@{}l@{}}Stlucia\\311417\end{tabular}       & \begin{tabular}[c]{@{}l@{}}~ Stlucia\\235510\end{tabular}                                  & \begin{tabular}[c]{@{}l@{}}\textbf{\%62}\\\textbf{\%58.9}\end{tabular}                     & \begin{tabular}[c]{@{}l@{}}\textbf{\%72}\\\textbf{\%99}\end{tabular}                 & \%55  & \%72                      \\ 
	\hline
	\begin{tabular}[c]{@{}l@{}}New College~\\160500\end{tabular}  & \begin{tabular}[c]{@{}l@{}}Newer College\\194000\end{tabular}                              & \begin{tabular}[c]{@{}l@{}}\textbf{Ground}\\\textbf{Truth}\end{tabular}                    & \begin{tabular}[c]{@{}l@{}}\textbf{Ground}\\\textbf{Truth}\end{tabular}              & ~\%44 & ~\%100                    \\
	\hline
\end{tabular}
\end{table}

\begin{figure*}[!htp]
\centering
\includegraphics[scale=0.4]{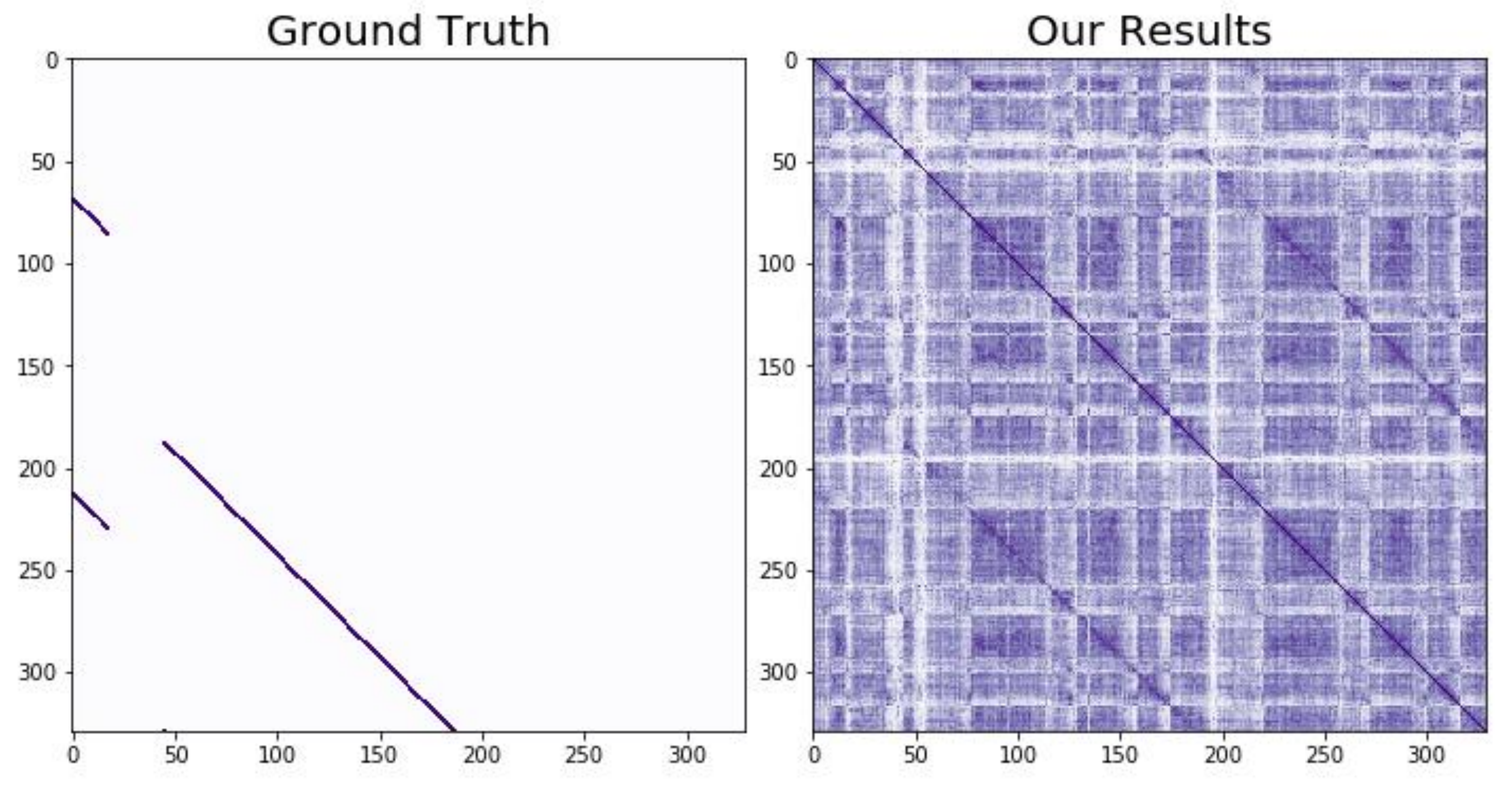}
\caption{ Ground Truth (left) and HGCN-BoW (right) confusion matrices for Lip6 indoor dataset dataset}\label{e}
\end{figure*}

Second row of the table  \ref{tab 1} shows the result of comparing four methods considering our training dataset is TUM sequence 11 and our testing dataset is dataset Lip6indoor lab.
The result of applying HGCN-BoW on dataset Lip6indoor dataset is plotted in Figure \ref{e}.
Here regular FABMAP2 and HGCN-FABMAP perform roughly the same, except at recall 81.5 in which HGCN-FABMAP
performs slightly better. Our conjecture behind this observation is that the environment TUM sequence 11 is not a perfect match for Lip6indoor dataset. Furthermore we have build the CLtree on a very small training set. \\ \\

\begin{figure*}[!htp]
\centering
\includegraphics[scale=0.4]{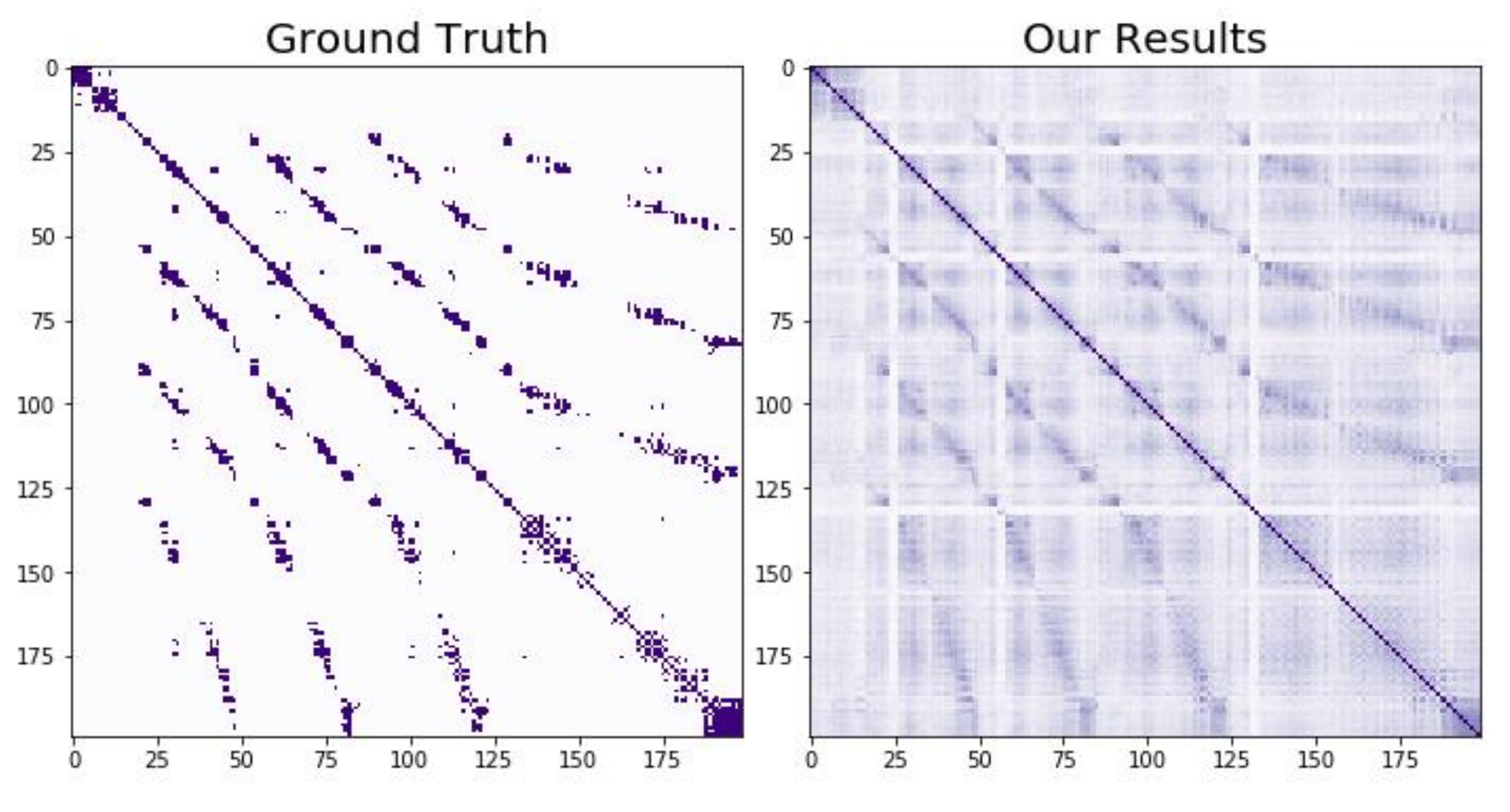}
\caption{Thresholded BoW (left), HGCN-BoW (right) for Stlucia Dataset}\label{f}
\end{figure*}

\begin{figure*}[!htp]
\centering
\includegraphics[scale=0.45]{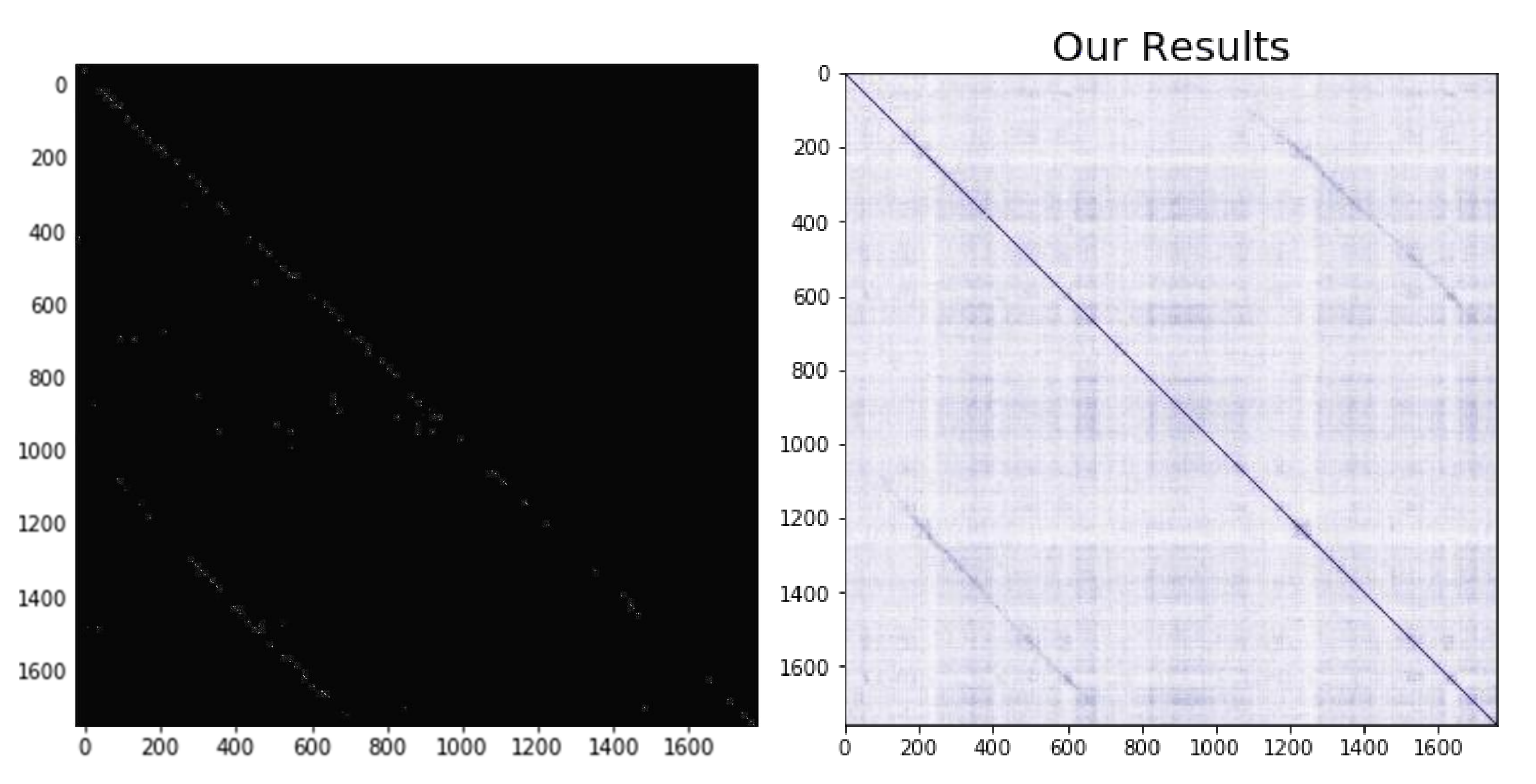}
\caption{FABMAP2 (left), HGCN-BoW (right) for Stlucia Dataset}\label{g}
\end{figure*}

The third row of table \ref{tab 1} shows that FABMAP2 and HGCN-FABMAP perform equally on Newer College dataset.
This dataset contains 3 loops. Figure \ref{f} shows that HGCN-FABMAP detects all of them. Due to lack of binary ground truth for this dataset, we have used a thresholded version of HGCN-FABMAP confusion matrix as our ground truth. 
The fourth row of  table \ref{tab 1} shows that HGCN-FABMAP outperforms regular FABMAP2, our justification is that as the number of SURF features in a tree like structure grows they become closer to each other therefore regular clustering algorithms fail to correctly identify the clusters, however when mapping those features to 
Poincaré ball this problem resolves due to the distances occurring between features. Figure \ref{g} shows that HGCN-BoW for  Stlucia dataset is a perfect match to ground truth.\\

Next we compare ORB-SLAM method with HGCN-ORB-SLAM, although we are not outperforming the algorithm but the results are competitive. Absolute RMSE for key frame trajectory (m) for  ORB-SLAM is  0.054353 and for our method it is 0.1.\\

\begin{figure*}[!htp]
\centering
\includegraphics[scale=0.9]{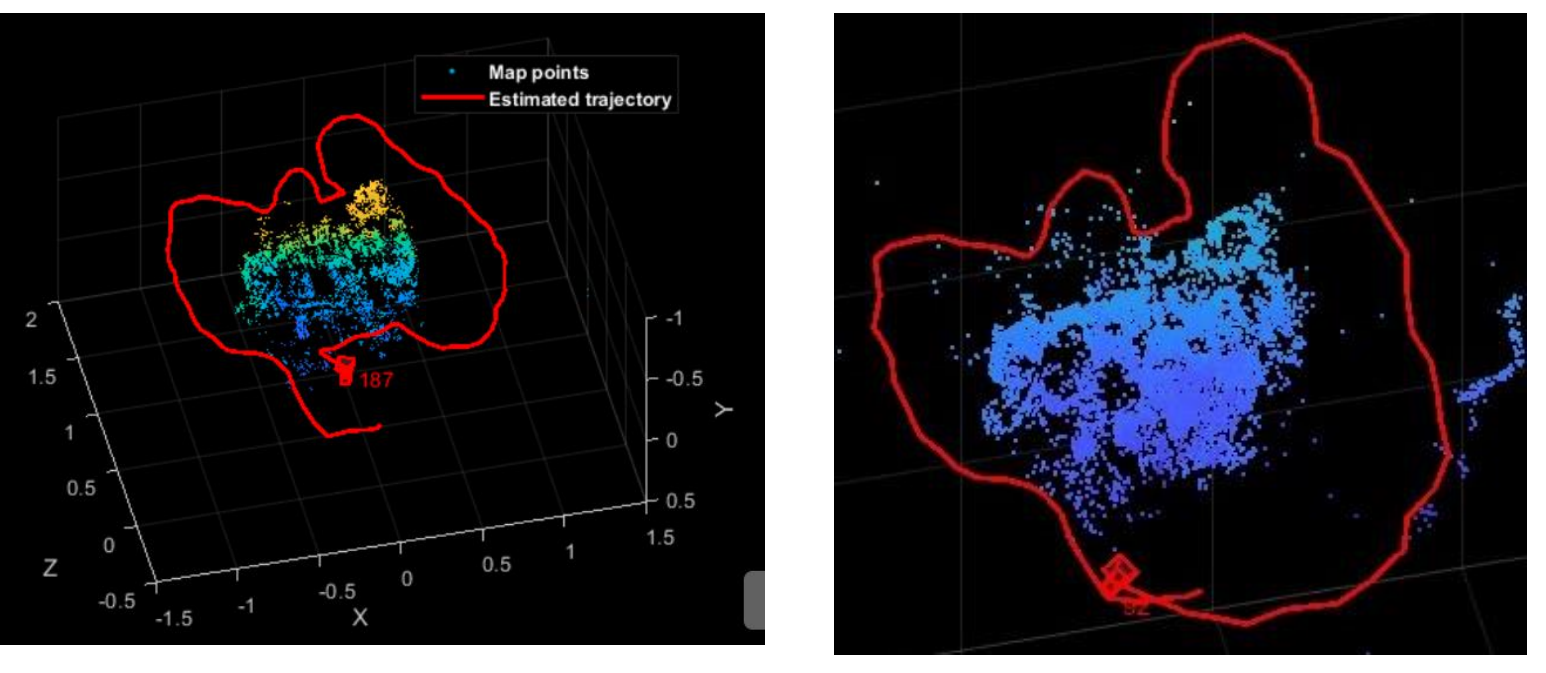}
\caption{ORB-SLAM(left), HGCN-ORB before applying optimization  camera trajectory)}\label{h}
\end{figure*}

To implement HGCN-ORB  we have modified the MATLAB package provided at the link \footnote[2]{https://www.mathworks.com/help/vision/ug/monocular-visual-simultaneous-localization-and-mapping.html}
Figure \ref{h} shows the comparison between these two methods.

\section{conclusion and future works}
In this paper we introduced extensions to the current state of the art SLAM methods, and compared them with FABMAP2, ORB-SLAM and BoW. We concluded that result of clustering in the vector quantization part of the SLAM may significantly enhance the loop closure detection accuracy and recall. Our next task is to extend HGCN-FABMAP to operate on long range. This  may not be achievable using hyperbolic graph convolutional neural networks because as the number of images increases in the number of features grows, and the problem becomes intractable using custom computers and servers. However through using semi-supervised Latent Dirichlet Allocation, (LDA), we can first compute local clusters for each dataset, and at the  second step join the resulted clusters to form  the global ones.

\nocite{*}
\bibliographystyle{ieeetr}
\bibliography{ms.bib}

\section*{Appendix}
\section*{Review of FABMAP1}
The following content is a brief tutorial about FABMAP model architecture, and it is adopted from  references,\cite{5509547} and  [\cite{5613942} - \cite{williams08iros}]
\subsection*{The FABMAP1 Model}
FABMAP2 learns a generative model the for bag of words data, based on the fact that, words co-occur with each other are usually  from the same  environment.\\

This system uses a bag of words representation of the received data, this representation  is chosen because of the decrease in learning and inference complexity it provides.
$Z_{k} = \{z_{1},...,z_{|\nu|}\}$, denotes a local scene observation at time $k$, where $z_i$ is a binary variable indicating the presence or absence of the $i^{th}$ word in the vocabulary. $Z^{k}$ represents the set of all observations up to time $k$.\\

At time $k$, the algorithm constructs a map from a set of locations $\mathcal{L}^{k} = \{L_{1},...,L_{n_{k}}\}$, each of which is associated with an appearance model.

The appearance model of a location is just a set of probabilities of scene elements exist at that location, i.e., $L_{i} = \{p(e_{i}=1|L_{i}),....,p(e_{|\nu|}=1|L_{i})\}$, where every $e_{i}$ is a scene elements that is generated independently by the location. The elements $e_q$ and the feature $z_q$ are linked using a detector model as follows:
 
\begin{center}
	$ \mathcal{D} = 
	\left\{
	\begin{array}{ll}
		p(z_{q}=1|e_{q}=0) \hspace{0.5cm} False \hspace{0.3cm} positive \hspace{0.3cm} probability \\
		p(z_{q}=0|e_{q}=1) \hspace{0.5cm} False \hspace{0.3cm} negative \hspace{0.3cm} probability 
	\end{array}
	\right.
	$
\end{center}

The purpose of introducing scene elements is to create a natural framework for combining data from many sources with various error formulations. Second, it allows us to split the $p(Z|L_{i})$ distribution into two sections. The first part is a simple model made up of variables called $e_q$. The second and more complicated part, which is built off-line and can be combined with a simple model based on the assumption that conditional dependencies between appearance words are independent of location and how this is achieved, explains the localization and mapping procedure through the use of Bayes filters.

Assume the robot is in the middle of the path and has captured some photographs, and we are only having a partial map. We calculate the likelihood of being at each position when the robot obtains new observations, assuming we have all of the observations acquired thus far.\\

\subsection*{Inference in FAB-MAP}
Calculating probability $p(L_{i}|\mathcal{Z}^{k})$ for each place $L_{i}$ is formulated as follow:\\
\begin{center}
	$ P(L_{i}|\mathcal{Z}^{k}) =\mathlarger{\frac{p(Z_{k}|L_{i},\mathcal{Z}^{k-1})p(L_{i}|\mathcal{Z}^{k-1})}{p(Z_{k}|\mathcal{Z}^{k-1})}}$ \\
\end{center}
in which $p(L_{i}|\mathcal{Z}^{k-1})$ is prior probability about the location. and $p(Z_{k}|L_{i},\mathcal{Z}^{k-1})$ is the observation likelihood and the denominator is the normalization term.

taking independence criteria between current and past observations into consideration, simplification of observation likelihood, $p(Z_{k}|L_{i},\mathcal{Z}^{k-1})$ leads to\\  \\
\begin{center}
$p(Z_{k}|L_{i}) = p(z_{n}|z_{1},...,z_{n-1},L_{i}) \times p(z_{n-1}|z_{1},...,z_{n-2},L_{i}) \times p(z_{2}|z_{1},L_{i}) \times p(z_{1}|L_{i})$ . \\
\end{center}

Due to high order conditional dependency between appearance words calculating
$p(Z_{k}|L_{i})$ is intractable, so this term is approximated using naive Bayes formula:\\
\begin{center}
	$p(Z_{k}|L_{i}) \approx \mathlarger{ \prod_{q=1}^{|\nu|} p(z_{q}|L_{i})}$
\end{center}
$p(z_{q}|L_{i})$ can be expanded like:\\
\begin{center}
	$p(z_{q}|L_{i}) = \sum_{s \in \{0,1\}} p(z_{q}|e_{q}=s,L_{i})p(e_{q}=s|L_{i})$.
\end{center}  
Furthermore, assuming errors are independent of position in the world $p(z_{q}|e_{q},L_{i}) = p(z_{q}|e_{q})$ therefore \\
\begin{center}
$p(z_{q}|L_{i}) = \sum_{s \in \{0,1\}} p(z_{q}|e_{q} =s)p(e_{q}=s|L_{i})$ \\
\end{center} For the sake of tractability we assume $p(e_{q}=S_{s_{q}}|z_{p_{q}},L_{i}) = p(e_{q}=s_{e_{q}}|L_{i})$, This yields:\\

\begin{center}
$p(z_{q}|z_{p_{q}},L_{i}) = \sum_{s_{e_{q}} \in \{0,1\}}p(z_{q}|e_{q}=s_{e_{q}},z_{p_{q}})p(e_{q}=s_{e_{q}}|L_{i})$
\end{center}

We need to compute $p(z^{k}|z^{k-1})$, which converts the appearance likelihood into a probability of loop closure, in order to get the probability that an observation is originating from somewhere not on the map. Our space must be partitioned into mapped and unmapped regions in order to calculate the $p(z^{k}|z^{k-1})$.

\begin{center}
$p(Z^{k}|Z^{k-1}) = \sum_{m \in \mathcal{L}^{k}}p(Z^{k}|L_{m})p(L_{m}|Z_{k-1}) + \sum_{u \in \mathcal{\overline{L}}^{k}}p(Z^{k}|L_{u})p(L_{u}|Z_{k-1})$.\\
\end{center}

 Unfortunately, the second term cannot be evaluated directly because it contains unknown locations, however using mean field approximation this term can be approximated.\\
 
\begin{center}
$p(Z_{k}|L_{avg}) =  \sum_{u \in \overline{\mathcal{L}^{k}}} p(L_{u}|Z_{k-1})$.\\
\end{center}

Where $\sum_{u \in \overline{\mathcal{L}^{k}}} p(L_{u}|Z_{k-1})$ is the prior probability of being in a new place and  
$p(Z_{k}|L_{avg})$ is the average place.

\end{document}